\documentclass[sigconf]{acmart}

\usepackage{amsmath}
\usepackage{mathtools}
\usepackage{amsthm}
\usepackage{caption}
\usepackage{microtype}
\usepackage{graphicx}
\usepackage{subfigure}
\usepackage{booktabs}
\usepackage{hyperref}
\usepackage{enumitem}
\usepackage{multirow}
\usepackage{multicol}
\usepackage{ulem}
\usepackage{algorithm}
\usepackage{algorithmic}

\usepackage{pifont}
\usepackage{xspace}

\usepackage{scalefnt}
\usepackage{wrapfig}
\usepackage{tabularx} 
\usepackage{colortbl}
\usepackage{xcolor,bm}
\usepackage{bm}

\definecolor{background_gray}{gray}{0.9}

\definecolor{mygreyr}{rgb}{0.99,0.985,0.98}
\definecolor{mygreyr1}{rgb}{0.98,0.975,0.97}
\definecolor{mygreyr2}{rgb}{0.97,0.965,0.96}
\definecolor{mygreyr3}{rgb}{0.96,0.955,0.95}
\definecolor{mygreyr4}{rgb}{0.955,0.95,0.945}
\definecolor{myred}{rgb}{0.995,0.9796,0.958}
\definecolor{myred1}{rgb}{0.992,0.9576,0.932}
\definecolor{myred2}{rgb}{0.991,0.9416,0.917}
\definecolor{myred3}{rgb}{0.990,0.9256,0.902}
\definecolor{myred4}{rgb}{0.990,0.915,0.89}

\AtBeginDocument{%
  }

\copyrightyear{2026}
\acmYear{2026}
\setcopyright{cc}
\setcctype{by}
\acmConference[KDD '26]{Proceedings of the 32nd ACM SIGKDD Conference on Knowledge Discovery and Data Mining V.1}{August 09--13, 2026}{Jeju Island, Republic of Korea}
\acmBooktitle{Proceedings of the 32nd ACM SIGKDD Conference on Knowledge Discovery and Data Mining V.1 (KDD '26), August 09--13, 2026, Jeju Island, Republic of Korea}
\acmPrice{}
\acmDOI{10.1145/3770854.3780193}
\acmISBN{979-8-4007-2258-5/2026/08}

\begin{document}

\title{Advanced Global Wildfire Activity Modeling with Hierarchical Graph ODE}
\newcommand{\method}{{\fontfamily{lmtt}\selectfont \textbf{HiGO}}\xspace}


\author{Fan Xu}
\affiliation{%
  \institution{University of Science and Technology of China}
  \city{Hefei}
  \state{Anhui}
  \country{China}
}
\email{markxu@mail.ustc.edu.cn}

\author{Wei Gong}
\affiliation{%
  \institution{University of Science and Technology of China}
  \city{Hefei}
  \state{Anhui}
  \country{China}
}
\email{weigong@ustc.edu.cn}

\author{Hao Wu}
\affiliation{%
  \institution{Tsinghua University}
  \city{Beijing}
  \country{China}
}
\email{wu-h25@mails.tsinghua.edu.cn}

\author{Lilan Peng}
\affiliation{%
  \institution{Southwest Jiaotong University}
  \city{Chengdu}
  \state{Sichuan}
  \country{China}
}
\email{llpeng@my.swjtu.edu.cn}

\author{Nan Wang}
\affiliation{%
  \institution{Beijing Jiaotong University}
  \city{Beijing}
  \country{China}
}
\email{wangnanbjtu@bjtu.edu.cn}

\author{Qingsong Wen}
\affiliation{%
  \institution{Squirrel AI}
  \city{Seattle}
  \state{Washington}
  \country{USA}
}
\email{qingsongedu@gmail.com}

\author{Xian Wu}
\authornote{Corresponding author. \\ Fan Xu is also with Shenzhen Loop Area Institute, Shenzhen, China.}
\affiliation{%
  \institution{Tencent}
  \city{Beijing}
  \country{China}
}
\email{kevinxwu@tencent.com}

\author{Kun Wang}
\affiliation{%
  \institution{Nanyang Technological University}
  \country{Singapore}
}
\email{wang.kun@ntu.edu.sg}

\author{Xibin Zhao}
\authornotemark[1]
\affiliation{%
  \institution{Tsinghua University}
  \city{Beijing}
  \country{China}
}
\email{zxb@tsinghua.edu.cn}

\renewcommand{\shortauthors}{Fan Xu et al.}


\begin{abstract}


Wildfires, as an integral component of the Earth system, are governed by a complex interplay of atmospheric, oceanic, and terrestrial processes spanning a vast range of spatiotemporal scales. Modeling their global activity on large timescales is therefore a critical yet challenging task. While deep learning has recently achieved significant breakthroughs in global weather forecasting, its potential for global wildfire behavior prediction remains underexplored. In this work, we reframe this problem and introduce the \textbf{\underline{Hi}}erarchical \textbf{\underline{G}}raph \textbf{\underline{O}}DE (\method{}), a novel framework designed to learn the multi-scale, continuous-time dynamics of wildfires. Specifically, we represent the Earth system as a multi-level graph hierarchy and propose an adaptive filtering message passing mechanism for both intra- and inter-level information flow, enabling more effective feature extraction and fusion. Furthermore, we incorporate GNN-parameterized Neural ODE modules at multiple levels to explicitly learn the continuous dynamics inherent to each scale. Through extensive experiments on the SeasFire Cube dataset, we demonstrate that \method{} significantly outperforms state-of-the-art baselines on long-range wildfire forecasting. Moreover, its continuous-time predictions exhibit strong observational consistency, highlighting its potential for real-world applications.


\end{abstract}




\maketitle

\newcommand\kddavailabilityurl{https://doi.org/10.5281/zenodo.18144805}
\ifdefempty{\kddavailabilityurl}{}{
\begingroup\small\noindent\raggedright\textbf{Resource Availability:}\\
The source code of this paper has been made publicly available at \url{\kddavailabilityurl}.
\endgroup
}

\section{Introduction}

\begin{figure}[t]
\centering
\includegraphics[width=0.98\linewidth]{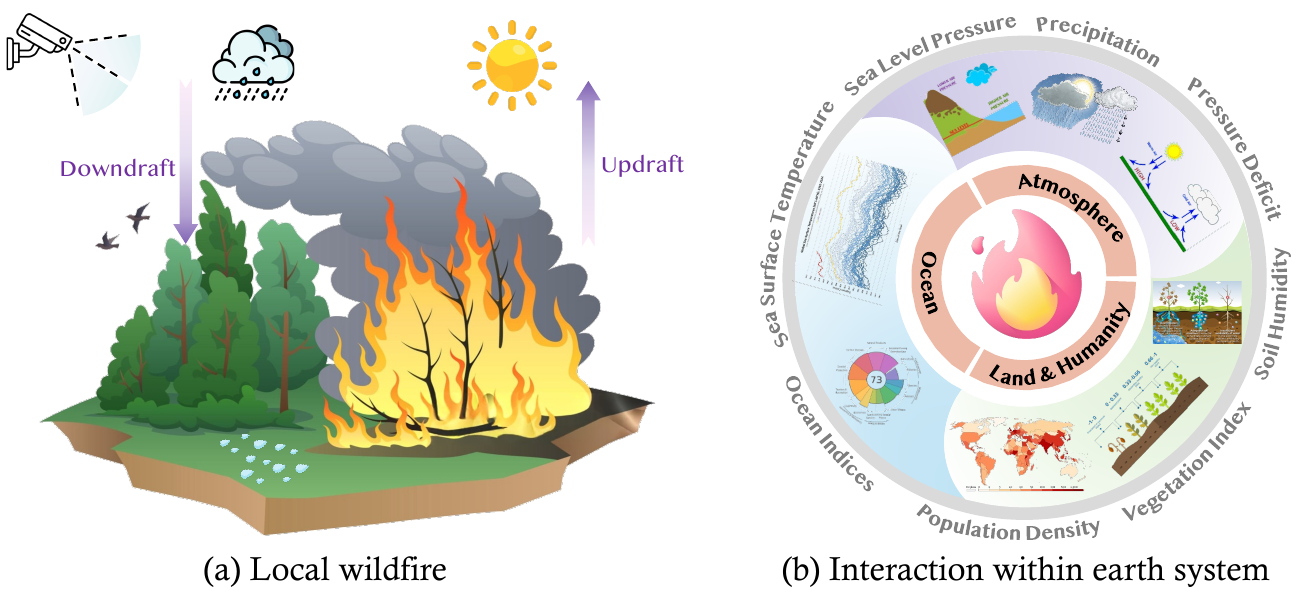}
\caption{The left depicts a local wildfire event, influenced by immediate conditions like airstreams. The right illustrates the broader Earth system context, framing wildfires as a phenomenon governed by a complex interaction between the Atmosphere, Ocean, Land and Human Activity.}
\label{intro}
\vspace{-5pt}
\end{figure}







Wildfires~\citep{sullivan2022spreading,bousfield2025global} are an integral component of the Earth system~\citep{flato2011earth,steffen2020emergence}, profoundly shaping global ecology and biogeochemical cycles~\citep{hedges1992global}. Driven by increasingly complex climate change, Wildfires' frequency, scale, and severity are intensifying at an unprecedented rate~\citep{brown2023climate}. This poses severe threats to human society and planetary health. Consequently, achieving accurate and reliable global wildfire activity forecast on large timescales has become a critical scientific pursuit.

As known, wildfires are not isolated local events but emergent phenomena resulting from complex, multi-scale interactions within the Earth system. Their evolution is governed by a hierarchy of processes: (1) global-scale climate modulations driven by ocean-atmosphere teleconnections~\citep{kucharski2010teleconnections}, such as the El Niño-Southern Oscillation (ENSO)~\citep{timmermann2018nino}; (2) regional-scale environmental shifts triggered by seasonal weather patterns~\citep{turrisi2021seasons}; and (3) local dynamics of fuel moisture and vegetation health~\citep{rodriguez2023modelling}. A successful global wildfire forecasting model must be able to capture and integrate these complex dependencies, which span vast spatiotemporal scales~\citep{cini2023scalable}. 

Driven by advancements in deep learning, its application within Earth system is rapidly evolving to accelerate scientific discovery~\citep{bi2022pangu,gao2025oneforecast}. In recent years, data-driven deep learning methods have achieved landmark success in domains such as weather forecasting and precipitation nowcasting~\citep{wu2025advanced}, where their performance is now highly competitive with state-of-the-art Numerical Weather Prediction (NWP)~\citep{brotzge2023challenges,chen2023fengwu} systems. However, the application of such methods to global wildfire forecasting has remained largely underexplored. 
The recent TeleViT~\citep{prapas2023televit} demonstrated the promise of Vision Transformers (ViTs)~\citep{dosovitskiy2020image} for modeling teleconnections, demonstrating the initial potential of deep learning for global wildfire modeling. However, the architecture's quadratic complexity with respect to the number of input tokens fundamentally limits its scalability~\citep{fei2025openck}. This creates a trade-off between predictive accuracy and computational efficiency, especially when modeling at high spatiotemporal resolutions. 

In our perspective, the researchers still face four key challenges in global wildfire activity forecasting: \ding{182} \textit{\textbf{Insufficient Problem Formulation:}} Most models oversimplify the task into a binary classification (\textit{fire vs. no-fire}), disregarding crucial quantitative information about local wildfire extent. \ding{183} \textit{\textbf{Suboptimal Feature Fusion:}} Lack of an explicit and effective mechanism to model the non-linear interactions and synergistic effects among heterogeneous driving factors from multiple physical domains (\textit{e.g.}, atmosphere, ocean, land)~\citep{karasante2025seasfire}. \ding{184} \textit{\textbf{Limited Feature Extraction Power:}} Existing methods struggle to simultaneously extract fine-grained features from high-resolution local data and capture long-range dependencies at a global scale. \ding{185} \textit{\textbf{Discrete-Time Modeling:}} Modeling the global worldfire evolution as discrete time steps~\citep{zang2020neural,chen2024signed} is inconsistent with the underlying continuous physical processes and can lead to the propagation and amplification of discretization errors over long-term forecasting. 

To address these challenges, we propose the \underline{\textbf{Hi}}erarchical \underline{\textbf{G}}raph \underline{\textbf{O}}DE (\method{}), a novel framework for global wildfire activity forecasting. Firstly, we reformulate the prediction task as an imbalanced, multi-class ordinal classification problem to capture wildfire activity with greater fidelity. Secondly, we incorporate multiple driver variables with the target variable with a refined channel attention and a cross-attention mechanism. Thirdly, we abstract the Earth system into a set of hierarchical graphs, where different levels capture physical interactions at varying granularities. We further design an adaptive attention-based message-passing mechanism to effectively learn the high-order representations of the inputs across the hierarchical structure. Finally, to capture the system's continuous dynamics, we integrate a neural ODE module parameterized by Graph Neural Networks (GNNs) at each hierarchical level. This design not only enables the model to learn the non-linear intrinsic evolutions but also naturally supports smooth predictions at any arbitrary time step, thereby enhancing the stability and physical consistency of long-range forecasting. We summarize the contributions of this paper as follows: 

\begin{itemize}[leftmargin=*]
    \item In this work, we reformulate the problem of global wildfire activity forecasting and introduce an effective data preprocessing and feature fusion pipeline.
    \item \method{} proposes a hierarchical graph structure and a novel adaptive message-passing architecture to efficiently capture high-order multi-scale dynamics.
    \item We integrate an advanced Graph ODE within the hierarchical graph structure for continuous-time modeling. Comprehensive experiments on the SeasFire Cube dataset demonstrate the superiority and effectiveness of our model.
\end{itemize}

\section{Related Work}

\subsection{Deep Learning with Earth System Modeling}

Deep learning has recently revolutionized Earth system science, particularly in weather forecasting. Foundational models such as Pangu-Weather~\citep{bi2022pangu}, FourCastNet~\citep{pathak2022fourcastnet}, and GraphCast~\citep{lam2023learning} have demonstrated that data-driven approaches can surpass traditional Numerical Weather Prediction (NWP) models~\citep{brotzge2023challenges,chen2023fengwu} in both accuracy and computational efficiency for medium-range forecasting. These models typically frame forecasting as a prediction task on a 2D grid, using architectures like U-Nets~\citep{ronneberger2015u}, Vision Transformers~\citep{dosovitskiy2020image, wu2024pastnet}, or GNNs~\citep{kipf2016semi} to predict future atmospheric states. Their success has inspired a shift towards applying similar data-driven paradigms to more complex, multi-domain problems like natural hazard forecasting~\citep{di2025global, wu2024pure}. 

For global wildfire activity modeling, this task presents unique challenges not fully addressed by the above weather forecasting models. Specifically, wildfire dynamics are governed by a slower, more complex interplay of atmospheric, oceanic, and terrestrial factors~\citep{jolly2015climate,stoof2025solve, gao2025neuralom}. Thus, our goal is to develop a novel modeling paradigm that explicitly captures these hybrid interactions and achieves stable forecasting.

\subsection{Spatiotemporal Graph Neural Networks}

To model complex, interacting systems defined on non-Euclidean domains, Graph Neural Networks (GNNs)~\citep{kipf2016semi} have emerged as a powerful and principled framework. GNNs operate on graph structures, learning representations of entities (nodes) by iteratively aggregating and transforming information from their local neighborhoods via a message-passing mechanism~\citep{wu2022graph}. 
In the context of Earth system science, the globe can be discretized into a grid, which naturally forms a graph where grid cells are nodes and spatial adjacency defines the edges~\citep{zhao2024beyond}. Each node can be endowed with a rich feature vector representing local physical states, such as temperature, fuel load, and moisture content, while edges can be weighted by interaction between nodes. Consequently, a GNN can learn to predict the evolution of node states (\textit{e.g.}, burning status, intensity) by explicitly modeling the spatiotemporal dependencies~\citep{yu2024deep} that govern the system's dynamics.

Prevailing spatio-temporal Graph Neural Networks~\citep{jin2023spatio,cini2023scalable}, such as STGCN~\citep{yu2017spatio} and DCRNN~\citep{li2017diffusion}, factorize the learning problem into distinct spatial and temporal modules. A GNN first extracts per-timestep spatial representations from the graph, which are then fed into a sequence model (\textit{e.g.}, GRU~\citep{qin2023hierarchically}, TCN~\citep{bai2018empirical}) to learn the temporal evolution. This decoupled design, while foundational, forms the basis for most standard benchmarks. However, the above formulations suffer from discrete-time restriction, while the evolution of wildfires is an inherently continuous-time process governed by underlying differential equations. Thus, in this work, we turn to combine the spatial modeling power of GNNs with the continuous-time modeling of Neural ODEs~\citep{chen2018neural}, namely graph ODE~\citep{fang2021spatial,chen2024signed, xu2025breaking}.

\section{Preliminary}

In this section, we delineate the foundational components of our study, including the data processing pipeline and the formal problem definition for global wildfire activity forecasting.

\subsection{Data Process of Wildfire and Driver Factor}

Our study is grounded in the \textit{SeasFire Cube} dataset~\citep{karasante2025seasfire}, a comprehensive, multivariate spatiotemporal repository designed for worldwide wildfire activity modeling. The dataset spans 21 years (2001-2021) with an 8-day temporal resolution. For our analysis, we leverage a curated set of variables representing key wildfire drivers and the target wildfire behavior. And all spatial data is coarsened from its original 0.25° resolution to a 1° grid to facilitate computationally efficient modeling of large-scale dynamics. This results in a global spatial resolution of size $H \times W = 180 \times 360$.

\subsubsection{Spatio-temporal Driver Variables.}
We select 10 driver variables that provide a multivariate snapshot of the Earth system's state, encompassing atmospheric, oceanic, terrestrial, and anthropic domains. These variables, henceforth denoted as the feature cube $\mathcal{X}$, are pre-processed to a uniform 1° spatial resolution. Specifically, for any given 8-day interval $t$, the state is represented by a tensor $\mathbf{X}_t \in \mathbb{R}^{H \times W \times C_x}$, where $C_x=10$ is the number of driver channels. The variables are: 
\ding{182} \textbf{Land}: Land Surface Temperature (day) ($lst\_day$), Normalized Difference Vegetation Index ($ndvi$), and Volumetric Soil Water layer 1 ($swvl1$).
\ding{183} \textbf{Anthropy}: Population Density ($pop\_dens$).
\ding{184} \textbf{Ocean}: Sea Surface Temperature ($sst$).
\ding{185} \textbf{Atmosphere}: Mean Sea Level Pressure ($mslp$), Surface Solar Radiation Downwards ($ssrd$), Mean 2m Temperature ($t2m\_mean$), Total Precipitation ($tp$), and Vapour Pressure Deficit ($vpd$).






Further, to mitigate the effects of highly skewed distributions, Total Precipitation ($tp$) and Population Density ($pop\_dens$) undergo a logarithmic transformation, i.e., $x' = \log(1+x)$.

\subsubsection{Climate Indice for Global Guidance.}
To incorporate long-range climate patterns (i.e., teleconnections) that are known to modulate regional fire regimes, we utilize 10 time-varying climate indices as an external global guidance. These indices, denoted as $\mathbf{z}_t \in \mathbb{R}^{C_z}$ ($C_z=10$), include Western Pacific Index, Pacific North American Index, North Atlantic Oscillation, Niño 3.4 Anomaly, \textit{et al}. More details can be found in Appendix~\ref{datadata}.

\subsubsection{Target Variable: Burned Area Quantization.}
The primary objective of our model is to forecast the global \textit{Burned Area} (BA), denoted by $\mathbf{Y}_t \in \mathbb{R}_{\ge 0}^{H \times W}$. The BA data is characterized by extreme sparsity and imbalance; at any given time step, wildfire activity typically covers only 1-2\% of the global land grid cells, with the vast majority of values being zero. While a binary classification (fire vs. no-fire) is feasible, it discards valuable information about fire severity and extent contained within the non-zero continuous values. Therefore, to capture a richer, more nuanced signal of wildfire behavior, we quantize the continuous BA values into a set of $K$ discrete, ordered levels $\{0, 1, ..., K-1\}$, where level 0 represents no fire, and levels $1$ to $K-1$ represent progressively increasing intensity or extent of fire. This transforms the forecasting task from a simple binary problem into a multi-class classification problem over a structured, ordinal output space. The resulting target tensor is $\mathbf{Y}_t \in \{0, 1, \dots, K-1\}^{H \times W}$. 
Throughout our study, the number of $K$ is fixed at 7 for both data preprocessing and model training.

\subsection{Problem Definition}

Following the established paradigm in data-driven Earth system forecasting, we formulate the global wildfire prediction task as a end-to-end formula, where the goal is to predict future states of the system based on its current condition. Specifically, our objective is to learn a mapping function $\mathcal{F}$ that takes the system's state at a given time $t$ and forecasts the global Burned Area (BA) pattern at a future time $t+\tau$, where $\tau$ is the forecast lead time or horizon.

Let $\mathbf{X}_t \in \mathbb{R}^{H \times W \times C_x}$ be the multivariate driver variables cube, $\mathbf{z}_t \in \mathbb{R}^{C_z}$ be the vector of climate indices, and $\mathbf{Y}_t \in \{0, 1, \dots, K-1\}^{H \times W}$ be the quantized Burned Area map at time step $t$. The forecasting model $\mathcal{F}$ aims to predict the target BA map $\mathbf{Y}_{t+\tau}$ based on the input state at time $t$. The formulation is as follows:
\begin{equation}
    \hat{\mathbf{Y}}_{t+\tau} = \mathcal{F}(\mathbf{X}_t, \mathbf{z}_t, \mathbf{Y}_t; \mathbf{\theta}),
\end{equation}
where $\hat{\mathbf{Y}}_{t+\tau}$ is the predicted BA map, and $\mathbf{\theta}$ represents the learnable parameters of the model $\mathcal{F}$. In our experimental setup, we investigate the model's performance across multiple forecasting horizons, $\tau \in \{h_1, h_2, \dots, h_N\}$, to assess its capabilities for both short-term (subseasonal) and long-term (seasonal) wildfire prediction.

\section{Methodology}

\begin{figure*}[t]
\centering
\includegraphics[width=1.0\linewidth]{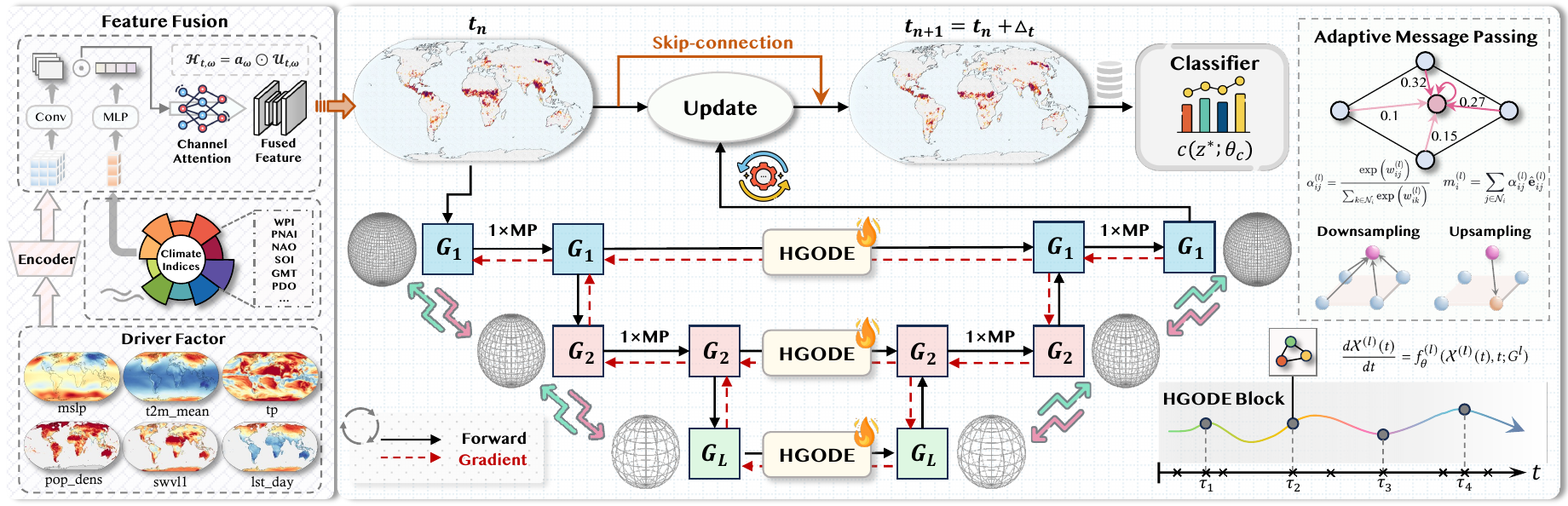}
\caption{Architectural overview of the HiGO framework. The framework integrates multi-modal driver and climate data to initialize a hierarchical graph representation. The core of the model consists of a multi-scale graph pyramid, where HGODE blocks learn continuous-time dynamics and an adaptive message-passing scheme captures complex spatial dependencies.}
\label{fig:framework}
\end{figure*}


\subsection{Overview}

Our proposed Hierarchical Graph ODE (\method{}) framework is designed to address the core challenges of global wildfire modeling by integrating principles from multi-scale graph representation, continuous-time dynamics, and multi-modal environmental variable fusion. The architecture is a fully differentiable, end-to-end pipeline that learns to predict global wildfire activity by capturing the complex, non-linear interactions across a hierarchy of spatio-temporal scales. Figure~\ref{fig:framework} provides a schematic overview of the \method{} framework, which comprises five main stages: (\romannumeral1) \textit{Climate-Informed Gated Feature Fusion}, (\romannumeral2) \textit{Hierarchical Graph Construction}, (\romannumeral3) \textit{Feature Mixer and Encoder}, (\romannumeral4) \textit{Learning Time-continuous Dynamics}, and (\romannumeral5) \textit{Decoder and Optimization}.

\subsection{Climate-Informed Gated Feature Fusion}

A formidable challenge in global wildfire activity modeling lies in the effective fusion of heterogeneous data modalities: the low-dimensional global climate indices that govern long-range interactions, and the high-dimensional local driver variables. To explicitly model the intricate, non-linear interactions between these two sources, we create a synergized input representation through an advanced channel attention module.

Let $\mathcal{V}_t \in \mathbb{R}^{H \times W \times C_x}$ denote the multi-channel tensor of spatio-temporal driver variables at timestep $t$, where $H, W$ are spatial dimensions and $C_x$ is the number of driver features (\textit{e.g.}, temperature, humidity). Let $\mathcal{I}_t \in \mathbb{R}^{C_z}$ be the corresponding vector of $C_z$ global climate indices (\textit{e.g.}, ENSO, NAO). 
First, we employ an MPL encoder $\phi_{\text{enc}}(\cdot)$ to project the driver variable tensor $\mathcal{V}_t$ into a latent embedding $\mathcal{Z}_t \in \mathbb{R}^{H \times W \times D}$.
Concurrently, we process the global climate index vector $\mathcal{I}_t \in \mathbb{R}^{C_z}$ to derive a set of high-level control signals. Inspired by the channel attention mechanism~\citep{takamoto2023learning}, we generate three distinct $d$-dimensional mask attention vectors $a_\omega~(\omega=1,2,3)$. Each vector is produced by a dedicated two-layer MLP:
\begin{equation}
    a_\omega=W_{2,\omega} \cdot \sigma(W_{1,\omega} \mathcal{I}_t+b_{1,\omega})+b_{2,\omega},
\end{equation}
where $W_{1/2,\omega}$ and $b_{1/2,\omega}$ are learnable weights, and $\sigma$ is the \text{GeLU} activation function. Intuitively, each vector $a_\omega$ can be conceptualized as assigning importance weights to different feature channels under a specific interpretation of the global climate context (\textit{e.g.}, an El Niño event might prompt the model to assign higher weights to drought-related channels).

To simulate the diverse physical dynamics, we apply a bank of three distinct convolutional operators $\kappa_\omega~(\omega=1,2,3)$ to the latent feature vector $\mathcal{Z}_t$, as follows: (1) 1 × 1 convolution ($\kappa_1$), (2) a depth-wise convolution ($\kappa_2$), (3) and a spectral convolution ($\kappa_3$). Each operator $\kappa_\omega$ transforms the latent representation $\mathcal{Z}_t$ to produce a "physically-processed" vector $\mathcal{U}_\omega \in \mathbb{R}^{H \times W \times D}$:
\begin{equation}
    \mathcal{U}_{t,\omega} = \kappa_\omega(\mathcal{Z}_t; \theta_{\kappa_\omega}).
\end{equation}
Then, we modulate these vectors with the climate-informed mask attention vectors $a_\omega$, and aggregate them to form the final fused feature representation $\mathcal{H}_t$.
\begin{equation}
    \mathcal{H}_{t, \omega}=a_{\omega} \odot \mathcal{U}_{t, \omega} ~ \Rightarrow ~ 
    \mathcal{H}_t = \mathcal{Z}_t + \sum_{\omega} \mathcal{H}_{t,\omega},
\end{equation}
where $\odot$ is the Hadamard operator over the channel dimension. Currently, we have performed comprehensive feature fusion with rich driver variables and global climate indices.

\subsection{Hierarchical Graph Construction}

To efficiently model long-range dependencies, we construct a multi-scale graph pyramid $\mathcal{G} = \{G_1, G_2, \dots, G_L\}$. The base graph $G_1 = (V_1, E_1)$ is derived from the input $H \times W$ grid, where nodes $V_1$ correspond to grid points and edges $E_1$ connect 4-way adjacent neighbors. For each level $l \in [1, L-1]$, we generate a coarser graph $G_{l+1}$ by applying a deterministic pooling operation on $G_l$. Specifically, each node $v_{i,j}^{(l+1)} \in V_{l+1}$ represents a non-overlapping $2 \times 2$ block of nodes from $V_l$, effectively halving the graph dimensions at each subsequent level. The intra-level edges $E_{l+1}$ are formed by again connecting 4-way neighbors within the coarser grid. To facilitate information flow between scales, we define inter-level edges $E_{l, l+1}$ connecting each node in $G_l$ to its corresponding parent node in $G_{l+1}$. The features for the intra-level and inter-level edges are initialized to a constant value of one. This predefined hierarchical structure provides a static, multi-resolutional structure for propagating physical dynamics across different spatial resolutions.

\subsection{Feature Mixer and Encoder}

Before conducting representation learning on the constructed hierarchical graphs, a critical step is to incorporate current state of the wildfire's Burning Area with the previously obtained driver-based fused representation $\mathcal{H}_t$. 

Let $\mathcal{B}_t \in \mathbb{R}^{H \times W \times 1}$ denote the single-channel tensor of the Burning Area at timestep $t$. To align its feature space with our fused representation, we first process $\mathcal{B}_t$ using a 1D convolutional layer, $\phi_{\text{ba}}(\cdot)$, which projects the burning features into a $D$-dimensional embedding $\hat{\mathcal{B}}_t$.
\begin{equation}
    \hat{\mathcal{B}}_t = \phi_{\text{ba}}(\mathcal{B}_t) \in \mathbb{R}^{H \times W \times D}.
\end{equation}
Then, we employ a cross-attention mechanism to explicitly model the interaction of the climate-informed driver representation and the current burning state. Specifically, we treat the fused feature representation $\mathcal{H}_t$ as the \textit{query} and the encoded burning embedding $\hat{\mathcal{B}}_t$ as both the \textit{key} and \textit{value}. The cross-attention operation~\citep{wei2020multi} is formulated as:
\begin{equation}
    \text{CrossAtt}(\mathcal{H}_t, \hat{\mathcal{B}}_t) = \text{softmax}\left(\frac{(\mathcal{H}_t W_Q)(\hat{\mathcal{B}}_t W_K)^T}{\sqrt{d_k}}\right)(\hat{\mathcal{B}}_t W_V).
\end{equation}
The learnable projection matrices $W_Q, W_K, W_V \in \mathbb{R}^{D \times D}$ map the inputs to the attention subspace, and $d_k$ is the dimension of the keys. The output of this operation, $\mathcal{A}_t \in \mathbb{R}^{H \times W \times D}$, is an attention-modulated representation of the burning state.
Further, we produce the \textit{initial state} representation for our dynamics model, $\mathcal{X}_t \in \mathbb{R}^{H \times W \times D}$, by combining the original fused features with the context-aware burning features via a residual connection.
\begin{equation}
    \mathcal{X}_t = \text{LayerNorm}(\text{FFN}( \mathcal{A}_t)) + \hat{\mathcal{B}}_t.
\end{equation}
Each vector at position $(i,j)$ in $\mathcal{X}_t$ is assigned as the feature vector for the corresponding node $v_{i,j}$, thus initializing the state on the hierarchical graph structure.

\subsection{Learning Time-continuous Dynamics}

To capture the continuous and complex non-linear evolution of global wildfire dynamics, we introduce a \textit{Hierarchical Graph ODE module} to operate over the hierarchical graph $\mathcal{G}$. The core of this stage is composed of adaptive message passing mechanism, inter-level propagation, and an advanced graph ODE.

\subsubsection{Adaptive Intra-Level Message Passing}

Firstly, we propose an Adaptive Message Passing (AdMP) mechanism to locally propagate information at each level of the hierarchical graph. 
Unlike isotropic aggregation methods that treat all neighbors uniformly, our AdMP learns to dynamically re-weight the contribution of each neighbor based on the current physical context. This is crucial for modeling directed phenomena such as environment-driven wildfire spread.

Given the node features $\mathbf{x}_i^{(l)}$ and edge features $\mathbf{e}_{ij}^{(l)}$ at level $l$, we first compute a scalar importance weight $w_{ij}^{(l)}$ for each edge, conditioned on the node and edge states.
\begin{equation}
    w_{ij}^{(l)} = \phi_{\text{edge}}^{(l)}(\mathbf{x}_i^{(l)}, \mathbf{x}_j^{(l)}, \mathbf{e}_{ij}^{(l)}),
\end{equation}
where $\phi_{\text{edge}}^{(l)}$ is a level-specific MLP. These importance weights are then normalized across the neighborhood $\mathcal{N}_i$ of node $i$ using a softmax function to produce the final attention coefficients $\alpha_{ij}^{(l)}$. Further, the aggregated message $\mathbf{m}_i^{(l)}$ for node $i$ is a weighted sum of the updated edge features from its neighbors. In formulation:
\begin{equation}
    \alpha_{ij}^{(l)} = \frac{\exp(w_{ij}^{(l)})}{\sum_{k \in \mathcal{N}_i} \exp(w_{ik}^{(l)})}, ~~~~
    m_i^{(l)} = \sum_{j \in \mathcal{N}_i} \alpha_{ij}^{(l)} \hat{\mathbf{e}}_{ij}^{(l)}.
\end{equation}
This adaptive aggregation allows the model to selectively focus on the most relevant incoming information.

\subsubsection{Learnable Inter-Level Propagation}

Effective hierarchical modeling requires a robust mechanism for exchanging information between hierarchical levels. We implement learnable downsampling and upsampling operations that leverage an attention-based strategy to preserve and refine information during transitions.

\noindent\paragraph{Downsampling.}
The downsampling operation projects information from a finer graph $G_l$ to a coarser graph $G_{l+1}$. For each node $v_i^{(l+1)} \in V_{l+1}$, its state is computed by aggregating information from its four corresponding children nodes in $V_l$, denoted by $\mathcal{C}(v_i^{(l+1)})$. We introduce a learnable attention mechanism where the attention coefficients are derived from the children's features. 
We denote the features of the children nodes as $\{\mathbf{x}_k^{(l)}\}_{k \in \mathcal{C}(v_i^{(l+1)})}$, and first compute inter-level attention coefficients $\beta_k^{(l)}$:
\begin{equation}
    \beta_k^{(l)} = \phi_{\text{down}}^{(l)}(\{\mathbf{x}_k^{(l)}\}_{k \in \mathcal{C}(v_i^{(l+1)})}),
\end{equation}
where $\phi_{\text{down}}^{(l)}$ is an MLP that takes the concatenated features of the four children as input and outputs four corresponding scalar weights, which are subsequently normalized via softmax. The state of the parent node $\mathbf{x}_i^{(l+1)}$ is then a weighted average of its children's states:
\begin{equation}
    \mathbf{x}_i^{(l+1)} = \sum_{k \in \mathcal{C}(v_i^{(l+1)})} \beta_k^{(l)} \mathbf{x}_k^{(l)}.
\end{equation}
\noindent\paragraph{Upsampling.}
The upsampling operation transfers information from a coarser graph $G_{l+1}$ back to the finer graph $G_l$. For each node $v_k^{(l)} \in V_l$, its upsampled feature is initialized with the feature of its parent node, $p(v_k^{(l)})$. To avoid the information loss and blocky artifacts associated with simple replication, we leverage the attention coefficients $\beta_k^{(l)}$ from the downsampling path to perform a skip connection. The final updated node feature $\tilde{\mathbf{x}}_k^{(l)}$ is formulated as a weighted combination:
\begin{equation}
    \tilde{\mathbf{x}}_k^{(l)} = \text{LayerNorm}\left( (1 - \beta_k^{(l)}) \cdot \mathbf{x}_k^{(l)} + \beta_k^{(l)} \cdot \mathbf{x}_{p(v_k^{(l)})}^{(l+1)} \right),
\end{equation}
where the pre-computed weight $\beta_k^{(l)}$ determines the blend between high-resolution local details and coarser-level context. This symmetric, attention-guided upsampling preserves feature fidelity more effectively than a generic message passing step.

\subsubsection{Hierarchical Graph ODE}

The temporal dynamics are modeled by defining a function $f_\theta$ that approximates the derivative of the system's state with respect to time, $d\mathcal{X}/dt = f_\theta(\mathcal{X}, t)$, where $\mathcal{X}$ represents the collective state of all nodes across all levels. 
At each level $l$, the dynamics function $f_\theta^{(l)}$ computes the state derivative based on the current node states at that level. The overall dynamics function is the collective operation across all levels:
\begin{equation}
    \frac{d\mathcal{X}^{(l)}(t)}{dt} = f_\theta^{(l)}(\mathcal{X}^{(l)}(t), t; G^{l}).
\end{equation}
The evolution of the system's state from time $t_0$ to $t_1$ is obtained by solving this system of ordinary differential equations using a numerical ODE solver, such as a Dormand-Prince or Runge-Kutta method~\citep{schober2014probabilistic}:
\begin{equation}
    \mathcal{X}(t_1) = \mathcal{X}(t_0) + \int_{t_0}^{t_1} f_\theta(\mathcal{X}(t), t) dt.
\end{equation}
This formulation allows for a continuous representation of the temporal dynamics, offering flexibility in prediction horizons and resilience to irregular time steps.

\subsection{Decoder and Optimization}

\paragraph{Decoder.}

The whole procedure contains intra-level message passing, intra-level continuous evolution modeling, and inter-level information propagation.
Ultimately, all learned spatio-temporal features are aggregated into this finest grid, creating a representation that is both locally precise and globally consistent. 
This is then passed through a final decoding MLP, $\phi_{\text{dec}}(\cdot)$, to produce the logit for wildfire occurrence at each spatial location. A sigmoid activation function, $\sigma(\cdot)$, is then applied to transform these logits into a probability map, $\hat{\mathcal{P}}_{t+1} \in [0, 1]^{H \times W}$:
\begin{equation}
    \hat{\mathcal{P}}_{t+1} = \sigma(\phi_{\text{dec}}(\mathcal{X}_{t+1}^{(1)})).
\end{equation}
Each element $(\hat{\mathcal{P}}_{t+1})_{i,j}$ represents the model's predicted probability that the grid $(i,j)$ will be part of a Burning Area at timestep $t+1$.

\paragraph{Optimization Objective.}


We frame the task of predicting wildfire activity as a point-wise multi-class classification problem. The ground truth Burning Area, $\mathcal{B}_{t+1}$, is discretized into $K$ categorical levels, representing distinct states such as 'no-burn', 'low-intensity burn', and 'high-intensity burn'. Let the target tensor as $\mathcal{Y}_{t+1} \in \{0, 1, \dots, K-1\}^{H \times W}$.
Wildfire prediction is characterized by a severe class imbalance, where 'no-burn' instances are dominant. We therefore adopt a weighted cross-entropy loss, where the weight assigned to each class is inversely proportional to its frequency in the training data. This ensures the optimization process appropriately penalizes errors on the crucial but rare active fire classes. The formulation is as follows:
\begin{equation}
    \mathcal{L} = -\frac{1}{H \times W} \sum_{i=1}^{H}\sum_{j=1}^{W} \sum_{k=0}^{K-1} w_k \cdot \mathbb{I}(y_{i,j}=k) \log(\hat{p}_{i,j,k}),
\end{equation}
where $y_{i,j}$ is the ground truth class label for cell $(i,j)$, $\hat{p}_{i,j,k}$ is the model's predicted probability for class $k$ at that cell, $\mathbb{I}(\cdot)$ is the indicator function, and $w_k$ is the weight assigned to class $k$. The class weights $\{w_k\}_{k=0}^{K-1}$ are typically set inversely proportional to their frequencies in the training dataset to ensure a balanced optimization landscape.


\section{Experiment}

In this section, we conduct comprehensive experiments to answer the following research questions ($\mathcal{RQ}$):

\begin{itemize}
    \item[$\mathcal{RQ}$1:] Does \method{} perform stable long-term and continuous-time extrapolation on global wildfire activity modeling?
    \item[$\mathcal{RQ}$2:] How sensitive is global wildfire activity modeling to diverse environmental forcing variables under \method{}?
    \item[$\mathcal{RQ}$3:] How do key design decisions, spanning module selection and hyperparameter tuning, impact model performance?
\end{itemize}

\subsection{Experimental Settings}
\noindent\textbf{Datasets.} 
We conduct our experiments on the \textit{SeasFire Cube} dataset, which provides 21 years (2001--2021) of global wildfire data at an 8-day temporal and 0.25° spatial resolution. The overall inputs consist of 10 spatiotemporal driver variables (\textit{e.g.}, Land Surface Temperature, NDVI) and 10 global climate indices (\textit{e.g.}, ENSO, NAO). The prediction target is the Burned Area (BA), which we formulate as a multi-class ordinal classification problem by quantizing it into $K=5$ severity levels. To ensure a robust evaluation on unseen future data, we enforce a strict chronological data split: training on years 2001--2018 and testing on 2019--2021. More details of data preprocessing can be found in section 3.1 and Appendix D.

\noindent\textbf{Baselines.} 
To comprehensively illustrate the effectiveness of our proposed \method{} framework, we conduct extensive comparisons against a diverse set of baselines spanning three major categories. 
\ding{95} First, choose \textit{\textbf{traditional point-wise methods}} such as MLP~\citep{taud2017multilayer} and XGBoost~\citep{chen2016xgboost}, which serve as fundamental benchmarks by making predictions for each grid independently.
\ding{95} Second, we evaluate against powerful \textit{\textbf{vision-based models}} that treat the global grid as an image, including U-Net~\citep{ronneberger2015u}, Vision Transformer (ViT)~\citep{dosovitskiy2020image} and Swin Transformer (SwinT)~\citep{liu2021swin}.
\ding{95} Third, we include several \textit{\textbf{graph-based models}}, from the foundational GCN~\citep{kipf2016semi} and the anisotropic model NDCN~\citep{zang2020neural} to the GNN-based excellent weather forecasting model GraphCast~\citep{lam2023learning}.

\noindent\textbf{Metrics.} 
Evaluating wildfire prediction is challenging due to a severe class imbalance, where "no-fire" instances dominate the dataset. To circumvent misleading metrics like accuracy, we adopt a robust evaluation protocol. For the purpose of evaluation, we consolidate all distinct fire severity levels into a single "fire" class, transforming the task into a binary problem of "fire" vs. "no-fire".
On this binary setup, we employ two standard metrics for imbalanced learning. We report the \textit{\textbf{Macro F1-Score (M-F1)}} for the positive "fire" class to assess the harmonic mean of precision and recall. Additionally, we use the \textit{\textbf{Area Under the Precision-Recall Curve (AUPRC)}}. As a threshold-independent measure, AUPRC provides a comprehensive evaluation of a model's ranking quality, which is especially critical when the positive class is rare.

\noindent\textbf{Implementations.} 
All models are implemented using PyTorch and PyTorch Geometric (PyG)~\citep{fey2019fast}, and trained on an NVIDIA A100 GPU. All models are optimized using the AdamW optimizer with an initial learning rate of $1 \times 10^{-4}$ and a weight decay of $1 \times 10^{-5}$. We employ a cosine annealing scheduler to adjust the learning rate over 100 training epochs with a batch size of 8. Model selection is based on the best validation performance on the AUPRC metric.
For our \method{} model, we set the number of graph hierarchy levels to $L=3$ and use a hidden dimension of $D=256$. The continuous dynamics are integrated using the `dopri5' adaptive-step ODE solver~\citep{chen2018neural}. For all baseline models, we replicate the architectures described in their original papers and adopt the hyperparameter settings directly from their officially released source code to ensure a faithful comparison. All input variables are standardized using normalization based on statistics from the training set.

\subsection{Main Results (RQ1)}

In this subsection, we first demonstrate \method{}'s superiority in wildfire activity forecasting by benchmarking it against a comprehensive suite of baselines across horizons up to 48 days. Then, we quantitatively validate its continuous-time flexibility through measuring its performance on unseen intermediate time points.

\subsubsection{Long-term Prediction}

\begin{table*}[!t]
\centering
\caption{Long-term forecasting performance comparison across multiple lead times. We report Macro F1-Score (M-F1) and Area Under the Precision-Recall Curve (AUPRC). The best-performing model in each column is highlighted in \textbf{bold}, and the second-best is \underline{underlined}. HiGO consistently demonstrates superior performance, with the advantage becoming more pronounced at longer horizons.}
\vspace{-10pt}
\label{tab:long_term}
\resizebox{\textwidth}{!}{%
\begin{tabular}{ll ccc ccc ccc ccc}
\toprule
& & \multicolumn{2}{c}{\textbf{8 days}} & \multicolumn{2}{c}{\textbf{16 days}} & \multicolumn{2}{c}{\textbf{24 days}} & \multicolumn{2}{c}{\textbf{32 days}} & \multicolumn{2}{c}{\textbf{40 days}} & \multicolumn{2}{c}{\textbf{48 days}} \\
\cmidrule(lr){3-4} \cmidrule(lr){5-6} \cmidrule(lr){7-8} \cmidrule(lr){9-10} \cmidrule(lr){11-12} \cmidrule(lr){13-14}
\textbf{Category} & \textbf{Model} & M-F1 & AUPRC & M-F1 & AUPRC & M-F1 & AUPRC & M-F1 & AUPRC & M-F1 & AUPRC & M-F1 & AUPRC \\
\midrule
\multirow{2}{*}{Point-wise} 
& MLP [9]      & 0.451 & 0.473 & 0.408 & 0.382 & 0.361 & 0.348 & 0.320 & 0.277 & 0.281 & 0.248 & 0.265 & 0.223 \\
& XGBoost [2]  & 0.439 & 0.485 & 0.429 & 0.432 & 0.355 & 0.377 & 0.312 & 0.320 & 0.303 & 0.291 & 0.288 & 0.266 \\
\midrule
\multirow{3}{*}{Vision-based} & U-Net [8] & 0.495 & 0.551 & 0.449 & 0.498 & 0.407 & 0.452 & 0.369 & 0.411 & 0.334 & 0.375 & 0.322 & 0.343 \\
& ViT [3] & 0.543 & 0.623 & 0.478 & 0.587 & 0.465 & 0.518 & 0.417 & 0.458 & 0.364 & 0.441 & 0.337 & 0.406 \\
& SwinT [7] & 0.561 & 0.627 & 0.527 & \underline{0.604} & 0.486 & 0.535 & 0.438 & 0.493 & 0.382 & 0.475 & 0.356 & 0.430 \\
\midrule
\multirow{3}{*}{Graph-based} & GCN [5] & 0.518 & 0.582 & 0.471 & 0.528 & 0.428 & 0.481 & 0.389 & 0.439 & 0.353 & 0.447 & 0.321 & 0.366 \\
& NDCN [10] & 0.540 & 0.611 & 0.496 & 0.560 & 0.455 & 0.514 & 0.417 & 0.473 & 0.381 & 0.465 & 0.367 & 0.411 \\
& GraphCast [6] & \underline{0.575} & \underline{0.631} & \underline{0.532} & 0.598 & \underline{0.491} & \underline{0.556} & \underline{0.453} & \underline{0.515} & \underline{0.417} & \underline{0.512} & \underline{0.384} & \underline{0.474} \\
\midrule
\rowcolor{mygreyr4}\textbf{Ours} & \textbf{HiGO} & \textbf{0.581} & \textbf{0.653} & \textbf{0.548} & \textbf{0.629} & \textbf{0.517} & \textbf{0.607} & \textbf{0.484} & \textbf{0.573} & \textbf{0.452} & \textbf{0.551} & \textbf{0.423} & \textbf{0.522} \\
\bottomrule
\end{tabular}
}
\vspace{-5pt}
\end{table*}

\begin{figure*}[t]
\centering
\includegraphics[width=1.0\linewidth]{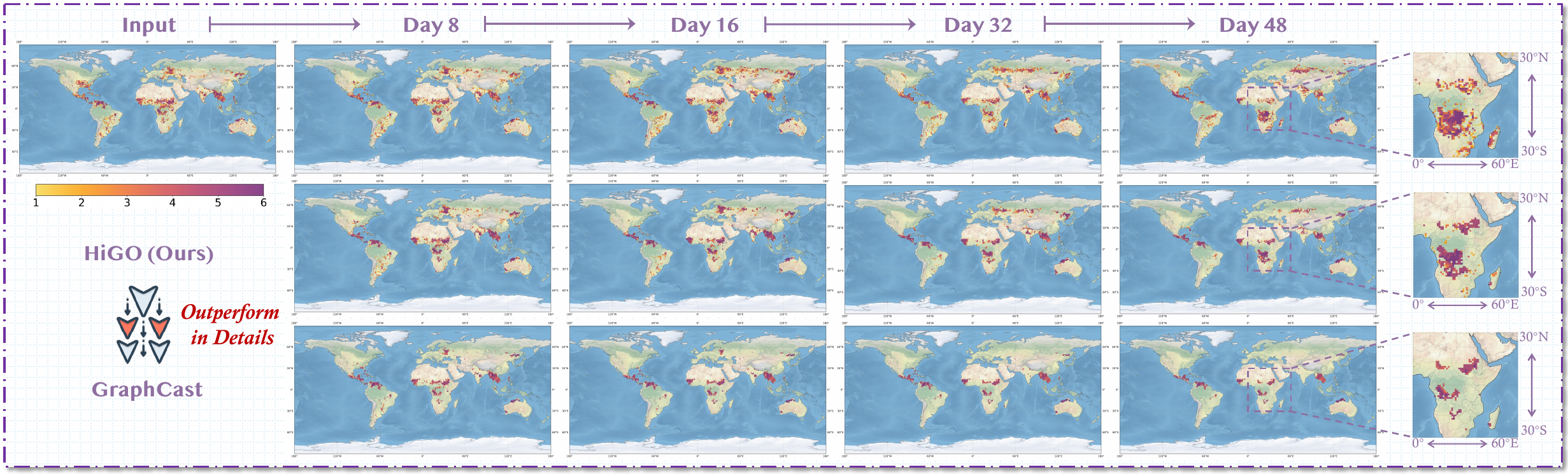}
\vspace{-17pt}
\caption{Visualization evaluation of HiGO's long-term wildfire activity forecasting performance. The panel provides a visual comparison between HiGO's multi-step predictions (top row) and the ground truth (bottom row) for up to a 48-day horizon, demonstrating high qualitative fidelity.}
\label{fig:exp1}
\vspace{-5pt}
\end{figure*}

To rigorously assess the long-term forecasting capabilities of our proposed model ($\mathcal{RQ}$1), we conduct a comprehensive evaluation against a suite of strong baselines across six distinct horizons, ranging from 8 to 48 days. The prediction target at each horizon is the quantized Burned Area, and the performance is measured using M-F1 and AUPRC metrics. The detailed results are presented in Table~\ref{tab:long_term}.

The results in Table~\ref{tab:long_term} reveal several key insights. First, as expected, point-wise methods (MLP, XGBoost) yield the lowest performance, as their inability to model spatial dependencies severely limits their predictive power. Second, vision-based models offer a substantial improvement by capturing local spatial patterns, with the standard SwinT outperforming U-Net and the standard ViT. Third, graph-based methods generally surpass vision-based approaches, validating our core hypothesis that representing the Earth system as a graph is more effective for modeling the complex, non-Euclidean interactions governing wildfire dynamics. Notably, GraphCast, which serves as a formidable baseline, almost consistently achieves the second-best results across most horizons. 

Our proposed \method{} model, however, establishes a new state of the art, demonstrating a consistent and significant performance advantage over all baselines across all lead times. The superiority of \method{} is particularly evident at longer horizons (\textit{e.g.}, 32-48 days). While the performance gap with GraphCast is modest at the 8-day lead time, it progressively widens as the forecast extends further into the future. For instance, at the 48-day mark, \method{} outperforms GraphCast by an absolute margin of 3.8\% on M-F1 and 4.1\% on AUPRC. We attribute this enhanced long-range stability to the stable graph ode module and the hierarchical graph structure, which enables the efficient fusion of information across multiple scales.
This quantitative advantage is further corroborated by the results in Figure~\ref{fig:exp1}. 
Crucially, at longer horizons (e.g., 48 days), GraphCast's forecasts tend to become overly diffuse (a likely artifact of error accumulation in discrete models). In contrast, HiGO's predictions remain sharp and physically plausible, visually demonstrating the stability afforded by its continuous-time formulation.

\subsubsection{Continuous-time Flexibility}

A central advantage claimed for our \method{}'s design is the inherent flexibility to predict at arbitrary time points. To provide a quantitative evaluation of this capability we conducted a targeted experiment designed to probe the model's performance at unseen intermediate time points. Specifically, we retrained the models on a fixed 16-day forecasting horizon, mapping the system state at time $t$ to $t+16$. This setup makes the ground-truth at the 8-day and 24-day serve as a rigorous testbed for evaluation. 

To ensure a comprehensive and fair comparison, we evaluate several modeling paradigms. We benchmark against GraphCast, a state-of-the-art auto-regressive graph model. Since it is inherently discrete and cannot predict at intermediate times, we use linear interpolation of its output probability maps to obtain results for the 8-day mark. For a direct comparison, we apply the same linear interpolation logic to our own model's 16-day output ("\method{} (Interp)"). Furthermore, to distinguish our contributions from the general concept of continuous-time dynamics, we also compare against NDCN, another graph-based model that utilizes a Neural ODE formulation. Our experimental results are presented in Table~\ref{tab:flexibility}. 

\begin{table}[t]
\centering
\caption{Quantitative evaluation of continuous-time flexibility. Models are trained on a 16-day horizon. Performance is then measured on an intermediate, unseen time-step (8-day \& 24-day)). Best results are in \textit{bold}.}
\vspace{-10pt}
\label{tab:flexibility}
\resizebox{0.9\columnwidth}{!}{
\begin{tabular}{l cc cc}
\toprule
& \multicolumn{2}{c}{\textbf{8-day}} & \multicolumn{2}{c}{\textbf{24-day}} \\
\cmidrule(lr){2-3} \cmidrule(lr){4-5}
\textbf{Model} & M-F1 & AUPRC & M-F1 & AUPRC \\
\midrule
GraphCast (Interp) & 0.484 & 0.492 & 0.459 & 0.512 \\
HiGO (Interp) & \underline{0.508} & 0.524 & 0.477 & 0.554 \\
NDCN (Conti) & 0.493 & \underline{0.537} & \underline{0.486} & \underline{0.561} \\
\rowcolor{mygreyr4}\textbf{HiGO (Conti)} & \textbf{0.551} & \textbf{0.623} & \textbf{0.507} & \textbf{0.597} \\
\bottomrule
\vspace{-10pt}
\end{tabular}
}
\end{table}

The results provide several key findings. First, simple linear interpolation does not work well for predicting intermediate wildfire states. This is true for both GraphCast and our model, \method{}, and it confirms that wildfire spread is a highly non-linear process. Second, \method{} performs significantly better than NDCN, another continuous-time model. While both models use a Neural ODE, the success of \method{} highlights the strength of its design. Its hierarchical graph structure and adaptive message-passing are crucial, allowing it to learn a more robust and physically realistic representation of the wildfire dynamics.

\subsection{Analysis of Forcing Variables (RQ2)}

To understand the contribution of different physical driver variables to our model's predictive performance, we conduct a comprehensive ablation study on the input forcing variables. We systematically evaluate the importance of each data modality by removing one group of variables at a time and observing the impact on long-range forecasting performance. 
We establish our full \method{} model, trained with all available inputs, as the performance baseline. We then train five experimental variants. Each variant omits exactly one of the following variable groups: (1) Land variables (\textit{e.g.}, vegetation index, soil moisture), (2) Ocean variables (sea surface temperature), (3) Atmosphere variables (\textit{e.g.}, temperature, pressure), (4) Anthropy variables (population density), and (5) all ten Climate Indices (\textit{e.g.}, ENSO, NAO). We evaluate these on the 24-day extrapolation task.

\begin{figure}[h]
\centering
\includegraphics[width=0.9\columnwidth]{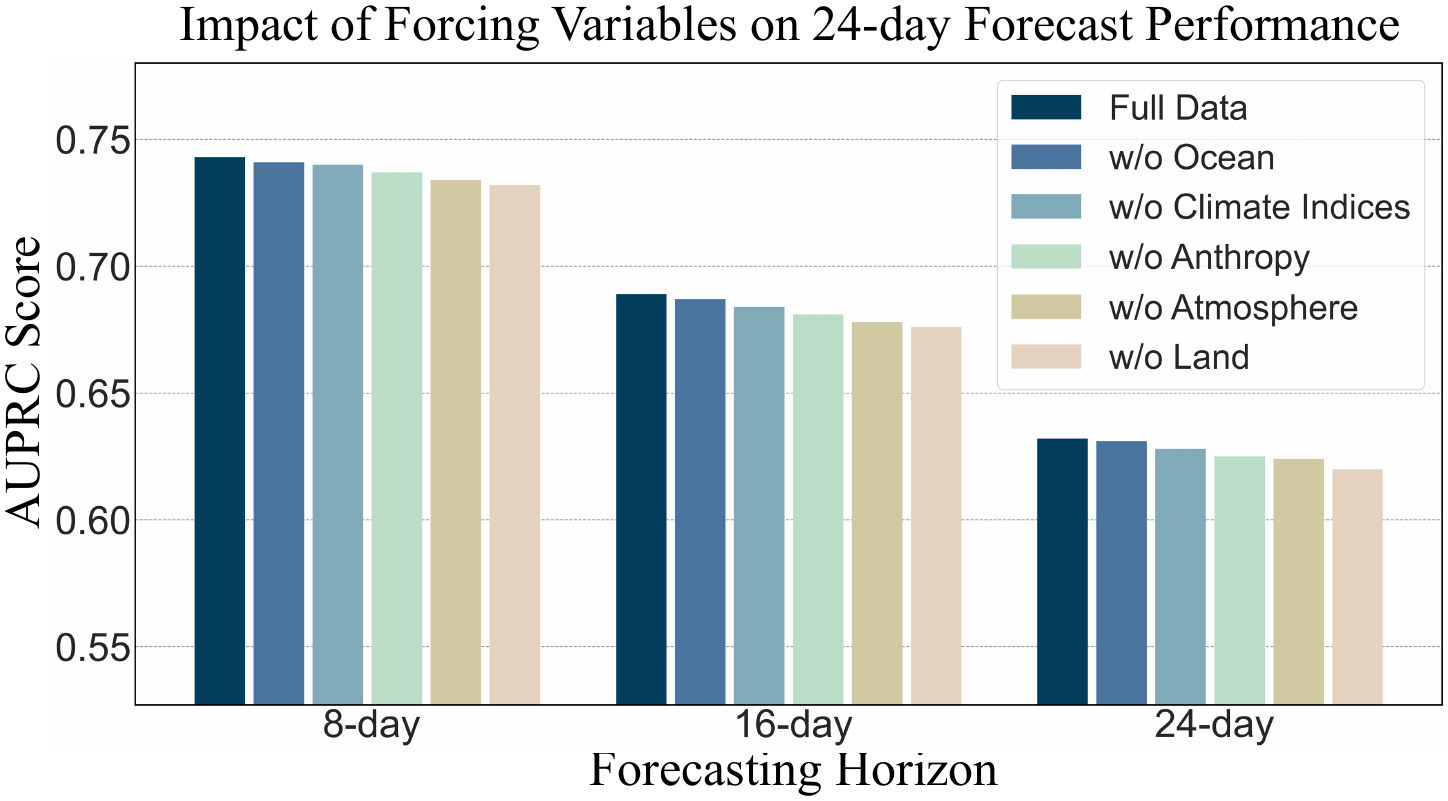} 
\vspace{-10pt}
\caption{Impact of Forcing Variables on 24-day Forecast Performance. The bar chart shows the AUPRC score for the full model (left) and five variants, each with one group of input variables removed.}
\label{fig:forcing_ablation}
\vspace{-5pt}
\end{figure}

The results are summarized in Figure~\ref{fig:forcing_ablation}, which presents a clear hierarchy of driver importance that remains consistent across all forecast horizons. Our analysis reveals that Land variables are the most critical factor, with their removal causing the most severe performance degradation. This underscores that terrestrial preconditions, such as fuel availability and soil moisture, are the dominant drivers of fire risk on long timescales. Atmospheric variables rank as the second most influential group, serving as key triggers for fire events. Conversely, the ablation of the other three driver groups (Ocean, Climate Indices, and Anthropy) leads to a less pronounced. For instance, the smaller impact of removing Ocean variables likely indicates that their primary influence on continental weather is already well-captured through the more proximal atmospheric variables and the broader climate indices. The results demonstrate that processing and integrating information from multi-domain variables is essential for building a robust and reliable wildfire forecasting system.

\subsection{Model Analysis (RQ4)}  

\subsubsection{Message-passing Mechanism}


To validate our adaptive attention-based message-passing design, we replace it with three standard graph filtering mechanisms while keeping the rest of the model architecture fixed. The variants include: \textbf{GCN}~\citep{kipf2016semi} (low-pass filter), \textbf{FAGCN}~\citep{bo2021beyond} (high-pass filter), and \textbf{BWGNN}~\citep{tang2022rethinking} (band-pass filter).

\begin{table}[t!]
\centering
\caption{Ablation study of the message-passing mechanism. Our adaptive approach is benchmarked against GNNs with fixed low-pass, high-pass, and band-pass filters on 16-day and 32-day forecast horizons. Best results are in bold.}
\vspace{-10pt}
\label{tab:message_passing_ablation}
\resizebox{1.0\columnwidth}{!}{%
\begin{tabular}{l l cc cc}
\toprule
& & \multicolumn{2}{c}{\textbf{16-day}} & \multicolumn{2}{c}{\textbf{32-day}} \\
\cmidrule(lr){3-4} \cmidrule(lr){5-6}
\textbf{Filter Type} & \textbf{Mechanism} & M-F1 & AUPRC & M-F1 & AUPRC \\
\midrule
Low-pass & GCN & 0.521 & 0.585 & 0.459 & 0.526 \\
High-pass & FAGCN & 0.530 & 0.598 & 0.468 & 0.534 \\
Band-pass & BWGNN & \underline{0.539} & \underline{0.607} & \underline{0.475} & \underline{0.542} \\
\midrule
\rowcolor{mygreyr4}\textbf{Adaptive (Ours)} & \textbf{\method{}} & \textbf{0.548} & \textbf{0.619} & \textbf{0.484} & \textbf{0.563} \\
\bottomrule
\end{tabular}
}
\vspace{-8pt} 
\end{table}

As shown in Table~\ref{tab:message_passing_ablation}, our adaptive approach consistently outperforms all fixed-filter variants across both 16-day and 32-day forecast horizons. The results indicate that a static filtering strategy, whether low-, high-, or band-pass, is insufficient for the complex dynamics of wildfire modeling. In contrast, our adaptive mechanism can learn to dynamically re-weight neighbor contributions, effectively acting as a context-aware filter. This flexibility is crucial for capturing the diverse, state-dependent interactions governing wildfire spread, leading to more accurate forecasts.

\subsubsection{Scale of Hierarchical Graph}

The hierarchical graph is a cornerstone of the \method{} framework, designed to capture multi-scale interactions within the Earth system. To investigate the impact of the hierarchy's depth ($\mathcal{RQ}$4), we vary the number of graph levels, $L$, from 1 (a single-scale graph) to 5, and evaluate the model's performance on both 8-day and 24-day forecasting horizons.

\begin{figure}[t]
\centering
\includegraphics[width=0.92\columnwidth]{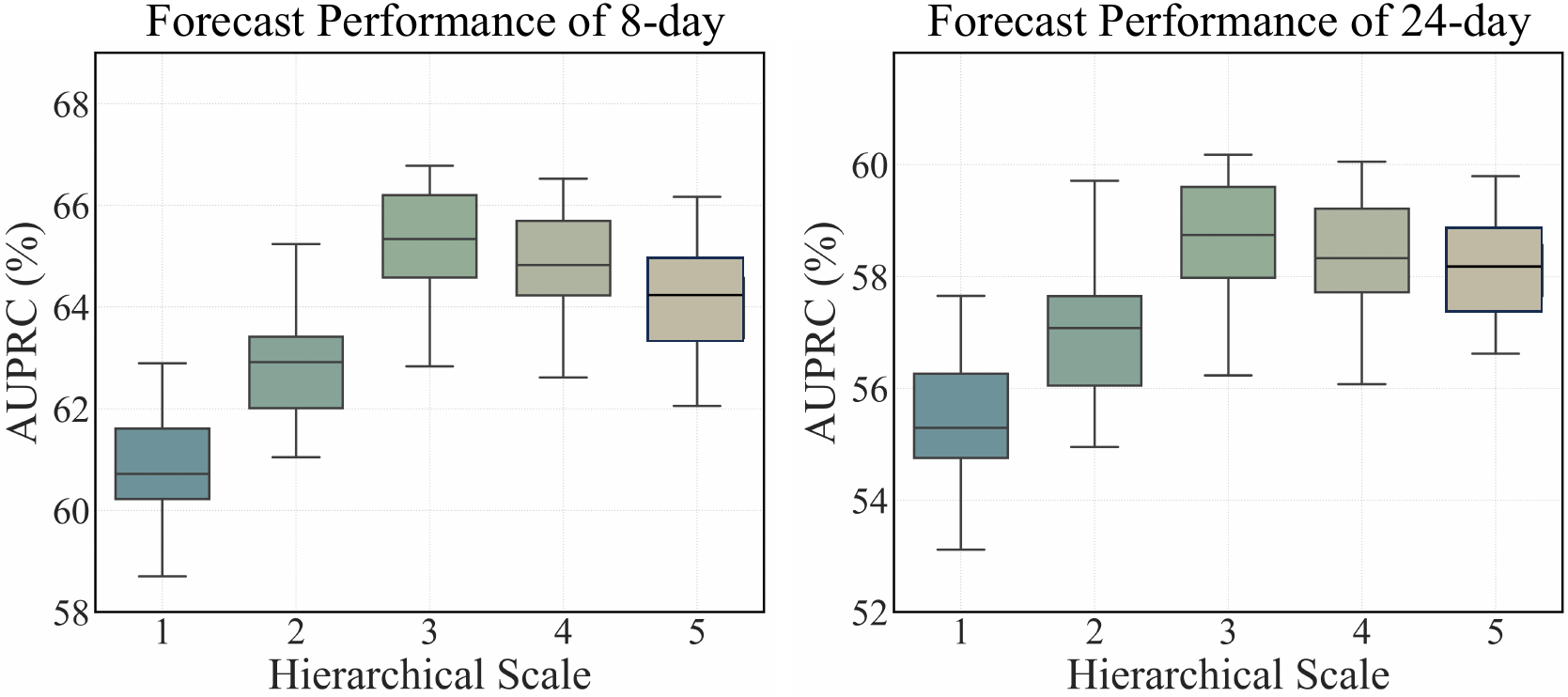} 
\vspace{-10pt}
\caption{Impact of the number of hierarchical graph levels ($L$) on forecasting performance.}
\label{fig:scale_ablation}
\vspace{-8pt}
\end{figure}

As illustrated in Figure~\ref{fig:scale_ablation}, the AUPRC performance initially improves as the number of levels increases from $L=1$ to $L=3$. The peak performance is achieved at $L=3$, which appears to strike an optimal balance between capturing sufficient multi-scale dynamics and preserving fine-grained details. However, further increasing the hierarchy depth to $L=4$ and $L=5$ yields no additional performance gains, and even shows a slight degradation. This suggests that excessively deep hierarchies introduce a longer and more complex information propagation path, potentially increasing the difficulty of optimization and leading to a form of over-smoothing across scales that erodes crucial fine-grained information.










\section{Conclusion}


In this work, we introduce \method{}, a novel framework designed for the complex task of global wildfire forecasting. Our approach is distinguished by its ability to effectively couple multi-source forcing data, such as large-scale climate indices and diverse driver variables, to guide forecasting. This is achieved through a sophisticated architecture that represents the Earth system as a multi-level graph hierarchy, capturing physical interactions across different scales. By integrating advanced Graph ODEs and an adaptive message-passing scheme, \method{} learns a physically-consistent, continuous-time representation of wildfire dynamics. Extensive experiments show that \method{} establishes state-of-the-art performance in long-range forecasting and continuous-time modeling in this scene.


\normalem  
\bibliographystyle{ACM-Reference-Format}
\bibliography{sample-base}

\appendix


\newpage
\appendix

\section{The Proposed \method{} Algorithm}

The pseudo-algorithm of \method{} is shown in Algorithm~\ref{alg:higo_training}.

\begin{algorithm}[h]
\caption{Training Procedure of \textsc{HiGO}}  
\label{alg:higo_training}
\begin{algorithmic}[1]
\REQUIRE Training data $\mathcal{D} = \{(\mathcal{V}_t, \mathcal{I}_t, \mathcal{B}_t, \mathcal{Y}_{t+\tau})\}$, learning rate $\eta$.
\STATE Initialize all model parameters $\boldsymbol{\Theta}$.
\FOR{number of training epochs}
    \STATE Sample a mini-batch $\{(\mathcal{V}, \mathcal{I}, \mathcal{B}, \mathcal{Y})\}$ from $\mathcal{D}$.
    
    \STATE \texttt{// Stage 1: Climate-Informed Gated Feature Fusion}
    \STATE Fused driver features $\mathcal{H}_t \leftarrow \text{ClimateGatedFusion}(\mathcal{V}, \mathcal{I})$.
    
    \STATE \texttt{// Stage 2: Feature Mixer and Encoder}
    \STATE Initial state $\mathcal{X}_t \leftarrow \text{FeatureMixer}(\mathcal{H}_t, \mathcal{B})$.
    
    \STATE \texttt{// Stage 3: Learning Time-continuous Dynamics}
    \STATE Final state $\mathcal{X}_{t+\tau} \leftarrow \text{HierarchicalGraphODE}(\mathcal{X}_t, [t, t+\tau])$.
    
    \STATE \texttt{// Stage 4: Decoder and Prediction}
    \STATE Predicted probabilities $\hat{\mathcal{P}}_{t+\tau} \leftarrow \text{Decoder}(\mathcal{X}_{t+\tau})$.
    
    \STATE \texttt{// Optimization}
    \STATE Compute weighted cross-entropy loss $\mathcal{L}(\hat{\mathcal{P}}_{t+\tau}, \mathcal{Y})$.
    \STATE Update parameters $\boldsymbol{\Theta}$ by descending the gradient: $\boldsymbol{\Theta} \leftarrow \boldsymbol{\Theta} - \eta \nabla_{\boldsymbol{\Theta}}\mathcal{L}$.
\ENDFOR
\RETURN Trained parameters $\boldsymbol{\Theta}^*$.
\end{algorithmic}
\end{algorithm}

\clearpage

\section{The Details of Input Data from SeasFire cube} \label{datadata}

The detailed description of diverse input variables and target variable are illustrated in Table~\ref{tab:input_variables}.

\begin{table}[h] 
\centering 
\caption{Input and target variables used from the SeasFire cube for all settings. The same variables are used for both local and global views.}
\vspace{-10pt}
\label{tab:input_variables}
\begin{tabular}{ll}
\toprule
\textbf{Full name} & \textbf{Pre-processing} \\
\midrule
\multicolumn{2}{l}{\textbf{Local/Global Variables}} \\
Mean sea level pressure & \\
Total precipitation & Log-transformed \\
Vapour Pressure Deficit & \\
Sea Surface Temperature & \\
Mean Temperature at 2 meters & \\
Surface solar radiation downwards & \\
Volumetric soil water level 1 & \\
Land Surface Temperature at day & \\
Normalized Difference Vegetation Index & \\
Population density & Log-transformed \\
\midrule
\multicolumn{2}{l}{\textbf{Climatic Indices}} \\
Western Pacific Index & \\
Pacific North American Index & \\
North Atlantic Oscillation & \\
Southern Oscillation Index & \\
Global Mean Temperature & \\
Pacific Decadal Oscillation & \\
Eastern Asia/Western Russia & \\
East Pacific/North Pacific Oscillation & \\
Niño 3.4 Anomaly & \\
Bivariate ENSO Timeseries & \\
\midrule
\multicolumn{2}{l}{\textbf{Target Variable}} \\
Burned Areas from GWIS & Multi-category \\
\bottomrule
\end{tabular}
\end{table}

\end{document}